\documentclass{article}
\usepackage{arxiv}
\usepackage[utf8]{inputenc}
\usepackage{graphicx}
\graphicspath{ {./images/} }
\usepackage{booktabs}
\usepackage{hyperref}

\title{Undivided Attention: Are Intermediate Layers Necessary for BERT?}

\author{Sharath Nittur Sridhar \\
    \small{Intel Labs, Intel Corporation} \\
    \small{sharath.nittur.sridhar@intel.com} \\\And
    Anthony Sarah \\
    \small{Intel Labs, Intel Corporation} \\
    \small{anthony.sarah@intel.com} \\\And
}

\date{}

\begin{document}

\maketitle

\begin{abstract}
In recent times, BERT-based models have been extremely successful in solving a variety of natural language processing (NLP) tasks such as reading comprehension, natural language inference, sentiment analysis, etc. All BERT-based architectures have a self-attention block followed by a block of intermediate layers as the basic building component. However, a strong justification for the inclusion of these intermediate layers remains missing in the literature. In this work we investigate the importance of intermediate layers on the overall network performance of downstream tasks. We show that reducing the number of intermediate layers and modifying the architecture for BERT\textsubscript{BASE} results in minimal loss in fine-tuning accuracy for downstream tasks while decreasing the number of parameters and training time of the model. Additionally, we use centered kernel alignment and probing linear classifiers to gain insight into our architectural modifications and justify that removal of intermediate layers has little impact on the fine-tuned accuracy.
\end{abstract}

\section{Introduction}

Language model pre-training has led to a number of breakthroughs in NLP (\cite{devlin2018bert}, \cite{brown2020language}, \cite{peters2018deep}, \cite{liu2019roberta} ) and have achieved state-of-the-art results on many non-trivial NLP tasks such as question-answering or natural language inference. In this problem domain, the BERT model, based on the \cite{vaswani2017attention} architecture has risen to prominence.

The BERT model architecture has multiple bidirectional Transformer (\cite{vaswani2017attention}) encoder blocks stacked together. Training BERT consists of first pre-training on large corpus of unlabeled data followed by fine-tuning on a much smaller, task-specific data set. Pre-training is computationally expensive and often requires several days to complete while fine-tuning can be completed in a much shorter time.

BERT models have significant parameter count and memory footprint, from 110M (BERT\textsubscript{BASE}) to 340M (BERT\textsubscript{LARGE}) to 8.3B (Megatron \cite{shoeybi2019megatron}) parameters. In this work, we demonstrate decreased model size and decreased training time with a negligible change in fine-tuning accuracy. While demonstrating decreased model size and decreased training time is relevant to the field in and of itself, we feel it is also important to a provide deeper understanding of why our architectural modifications are effective. Similar to the work done in \cite{clark2019does} we use probing linear classifiers and centered kernel alignment (\cite{kornblith2019similarity}) to gain insight into the improvements made by our architectural modifications.

\section{Related Work}

Approaches such as \cite{sanh2020distilbert},  \cite{sun2020mobilebert}, \cite{jiao2020tinybert}, \cite{sun2019patient} and \cite{xu2020bertoftheseus} use knowledge distillation techniques to reduce the parameter count in BERT and improve training speed. The authors of DistilBERT (\cite{sanh2020distilbert}) decreased the model size by 40\% and improved the training speed by 60\%, with minimal loss in accuracy. Similarly, the authors of TinyBERT (\cite{jiao2020tinybert}) employ a two-stage distillation framework, that performs distillation on both the pre-training and task specific learning stage to decrease the model size by 7.5x. BERT-PKD (Patient Knowledge Distillation) (\cite{sun2019patient}) learns from multiple layers of the teacher model for incremental knowledge extraction,  whereas the BERT-of-Theseus approach (\cite{xu2020bertoftheseus}) progressively substitutes modules of BERT with modules having fewer parameters. However, all knowledge distillation techniques require a pre-trained teacher model, which may not available in many cases.  

 Other approaches, such as ALBERT (\cite{lan2020albert}), employ a number of techniques, such as decomposing the vocabulary matrix and cross-layer parameter sharing, to achieve similar accuracy to BERT but with 18x fewer parameters and 1.7x faster training. The authors of \cite{press2020improving} changed the ordering of the sub-layers to create the \textit{sandwich transformer}. Like us, the authors recognize that the ordering of the sub-layers in BERT networks is not well justified nor necessarily optimal. However, unlike ALBERT, their goal is to improve language modeling performance and not to decrease the network size or complexity. In this work, we share similar goals to ALBERT but employ a different approach that removes blocks of intermediate layers from the network.

\section{Architectural Modifications}
\begin{figure}[h]
    \center{\includegraphics[scale=0.4] {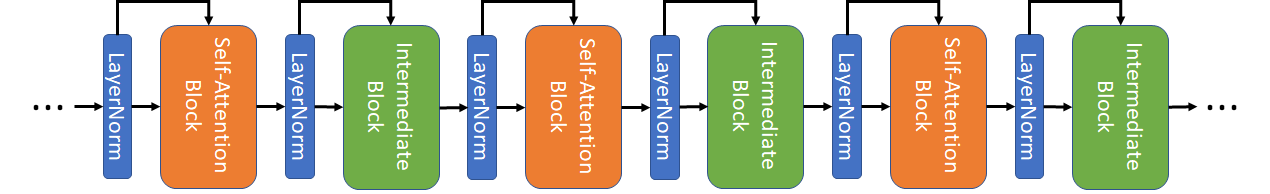}}
    \caption{\label{fig:unmodified_network}First six blocks of the unmodified network architecture. This BERT\textsubscript{BASE} network will have a total of 12 self-attention blocks and 12 intermediate blocks.}
\end{figure}

BERT \cite{devlin2018bert} is a stack of multiple Transformer encoder blocks with each encoder block further consisting of a separate multi-head self-attention block followed by an intermediate block (Figure \ref{fig:unmodified_network}). The importance of the self-attention block has been extensively analyzed in \cite{clark2019does}, \cite{michel2019sixteen}, \cite{tenney2019bert} and others.

\begin{figure}[h]
    \center{\includegraphics[scale=0.5] {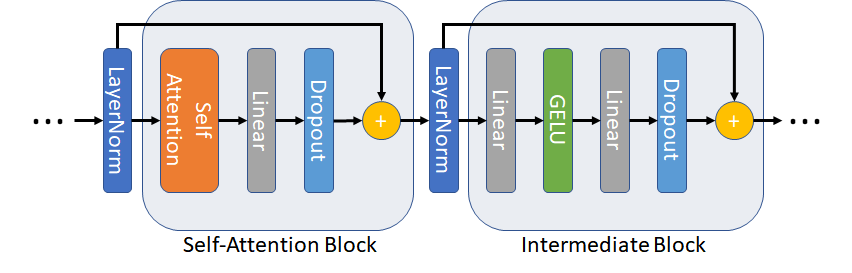}}
    \caption{\label{fig:reference_unit}Reference unit in the unmodified BERT network. The unmodified BERT\textsubscript{BASE} network contains 12 of these units arranged sequentially.}
\end{figure}

An intermediate block is a four-layer feed-forward network containing a Gaussian error linear unit (GELU) (\cite{hendrycks2020gaussian}) in between two linear layers followed by a dropout layer (see \textit{Intermediate Block} of Figure \ref{fig:reference_unit}). The intermediate block was added mainly to enrich the representations obtained from the self-attention block. However, the relevance of this block has not been well studied and no strong justification for their inclusion in Transformer-type architectures has been made in the literature.

\begin{figure}
    \center{\includegraphics[scale=0.4] {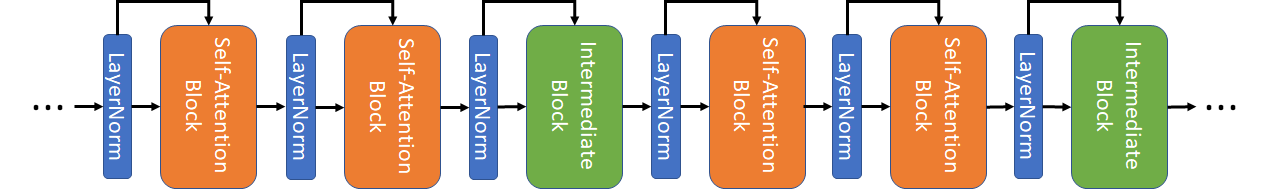}}
    \caption{\label{fig:modified_network}First six blocks of the modified network architecture with $n$ = 2. Note the removal of intermediate blocks between every $n$ = 2 self-attention blocks (compare to Figure \ref{fig:unmodified_network}). Using this architecture, the modified BERT\textsubscript{BASE} network will have only six intermediate blocks instead of the 12 which are found in the unmodified network.}
\end{figure}

Our main motivation is to decrease model size and complexity while quantifying and understanding any negative impact of reducing the number of intermediate blocks on model accuracy. To this end, we modify the BERT architecture by removing some of the intermediate blocks within the network. More specifically, an intermediate block will be added only after every $n$ self-attention blocks. If the total number of self-attention blocks in the network is $m$ then the modified network will contain $\lfloor\frac{m}{n}\rfloor$ intermediate blocks. For example, when $n$ = 1 the network is unmodified and contains $m$ intermediate blocks. However, when $n$ = $\infty$ the modified network contains no intermediate blocks. See Figure \ref{fig:modified_network} for an example of the modified BERT\textsubscript{BASE} network with $n$ = 2.

Note that $n$ is an architectural hyper-parameter that can be changed to make trade-offs in network size, complexity and accuracy. We experiment with different values of $n$ and analyze its effects with multiple fine-tuning tasks. 

\section{Results}
\label{section:convergence_results}
In this section we present the convergence and accuracy results for $n \in \{1,2,3,4,6,\infty\}$ where $n$ = 1 represents the unmodified network and $n$ = $\infty$ represents the modified network with no intermediate blocks. We chose BERT\textsubscript{BASE} as our reference model which has 12 self-attention blocks each of which is followed by an intermediate block with a hidden size of 768. We perform all our experiments on BERT\textsubscript{BASE}, mainly due to limited computational resources and the extremely long times required to train larger models. Table \ref{tab:combined_fine_tuning_results_parameters} shows the results of fine-tuning on downstream tasks along with the total number of parameters and the relative decrease in size and relative increase in throughput (measured as examples/sec) of the unmodified and modified BERT\textsubscript{BASE} networks. 

\begin{table}[h]
    \footnotesize
    \begin{center}
        \caption{Fine-tuning results for unmodified and modified BERT\textsubscript{BASE} networks with different values of $n$ along with the number of parameters, decrease in size and increase in throughput. The F1 and EM scores are shown for SQuAD v1.1, m/mm for MNLI, accuracy for SST-2 and F1 score for QQP. We compare our approach against  DistilBERT (\cite{sanh2020distilbert}), BERT-PKD (\cite{sun2019patient}) and BERT-of-Theseus (\cite{xu2020bertoftheseus}). For DistilBERT, we show the published scores and parameter count reported in  \cite{sanh2020distilbert}. Similarly, for the BERT-PKD and BERT-of-Theseus networks, we show the values reported in \cite{xu2020bertoftheseus}. Note that certain fine-tuning scores for DistilBERT, BERT-PKD and BERT-of-Theseus are missing in the table as they were not available in the published literature.  We do not show the throughput increase for these networks due to the difference in hardware platforms. }
        \label{tab:combined_fine_tuning_results_parameters}
        \begin{tabular}{c|c|c|c|c|c|c|c}
            \hline \hline
             \textbf{Network} & \textbf{SQuAD} & \textbf{MNLI} & \textbf{SST-2} & \textbf{QQP}  & \begin{tabular}{@{}c@{}}\textbf{Parameter} \\ \textbf{Count}\end{tabular} & \begin{tabular}{@{}c@{}}\textbf{Size} \\ \textbf{Decrease}\end{tabular} & \begin{tabular}{@{}c@{}}\textbf{Throughput} \\ \textbf{Increase}\end{tabular} \\
            \hline
            $n$ = 1 & 88.43/80.97 & 82.83/83.38 & 91.97 & 87.50 & 110.10M & 1.00x & 1.00x  \\
            $n$ = 2 & 87.09/79.66 & 81.64/82.32 & 89.44 & 87.38 & 81.76M & 1.35x & 1.39x \\
            $n$ = 3 &  86.91/79.28 & 81.41/82.20 & 90.13 & 87.13 & 72.31M & 1.52x & 1.59x  \\
            $n$ = 4 & 86.63/78.86 & 81.47/82.45 & 90.71 & 86.76 & 67.59M & 1.63x & 1.72x \\
            $n$ = 6 & 85.73/77.59 & 80.65/80.85 & 88.64 & 86.42 & 62.86M & 1.75x & 1.87x  \\
            $n$ = $\infty$ & 83.27/74.59 & 79.07/79.85 & 89.33 & 85.44 & 53.41M & 2.06x & 2.28x   \\
            DistilBERT & 85.8/77.7 & 82.2/- & 91.3 & - & 66M & 1.66x & -  \\
            BERT-PKD & -/- & 81.3/- & 91.3 & - & 66M & 1.66x & - \\ 
            BERT-of-Theseus & -/- & 82.3/- & 91.5 & - & 66M & 1.66x & - \\
        
            \hline \hline
        \end{tabular}
    \end{center}
\end{table}

\subsection{Pre-Training} \label{section:pre_training}
All variants were pre-trained from scratch on the English Wikipedia and BookCorpus (\cite{zhu2015aligning}) data sets. Similar to \cite{devlin2018bert}, we follow a two-phase pre-training approach. For the first 90\% of iterations, the model is trained with a sequence length of 128 and the remaining 10\% is trained with a sequence length of 512. To avoid having to search for optimal training hyper-parameters, we chose to use the same values for all variants. An initial learning rate of $10^{-4}$ was used for all variants including a linear warm-up schedule for the first 1\% of training.

\subsection{Fine-Tuning} \label{section:fine_tuning}
During fine-tuning, all weights of each network investigated are modified through training with a task-specific data set. We evaluate our models on two popular benchmarks: The General Language Understanding Evaluation (GLUE) benchmark \cite{wang2019glue} and the Stanford Question Answering Dataset (SQuAD) \cite{Rajpurkar_2016}. From the GLUE benchmark, we chose MNLI, SST-2, and QQP as our primary tasks since they have sufficiently large data sets and provide stable results across runs. All fine-tuning results are shown on the development sets of SQuAD, MNLI, SST-2 and QQP. 

From Table \ref{tab:combined_fine_tuning_results_parameters} we see that the modified BERT\textsubscript{BASE} networks perform quite well compared to the unmodified network. Specifically, with $n \in \{2, 3, 4\}$ there is approximately 1-2\% loss in accuracy across fine-tuning tasks while simultaneously providing a significant improvement in size and throughput (see Table \ref{tab:combined_fine_tuning_results_parameters}). These results indicate that the $n$ hyper-parameter can be used to make trade-offs in network size, speed and accuracy. For example, if an F1 score on SQuAD of approximately 87\% is acceptable then a modified ($n$ = 3) network could be used which would be approximately 1.5x smaller and 1.6x faster. Other trade-offs can be made for situations where the network is to be deployed onto a system with very limited memory and computational resources. In that case, minimizing network size and computational complexity would be critical and choosing a modified ($n$ = $\infty$) network would decrease memory usage by more than 2x and be approximately 2.3x faster than the unmodified ($n$ = 1) network.

In addition, the specific \textit{location} of any one intermediate block in the modified BERT\textsubscript{BASE} networks does not seem to have a disproportionate affect on performance. For example, with $n$ = 2, half of all intermediate blocks are removed from multiple locations within in the network and this results in only a slight accuracy degradation. Thus, the removal of any \textit{one} intermediate block, regardless of its location, would have an even smaller accuracy loss.

\begin{figure}[h]
    \center{\includegraphics[scale=0.3] {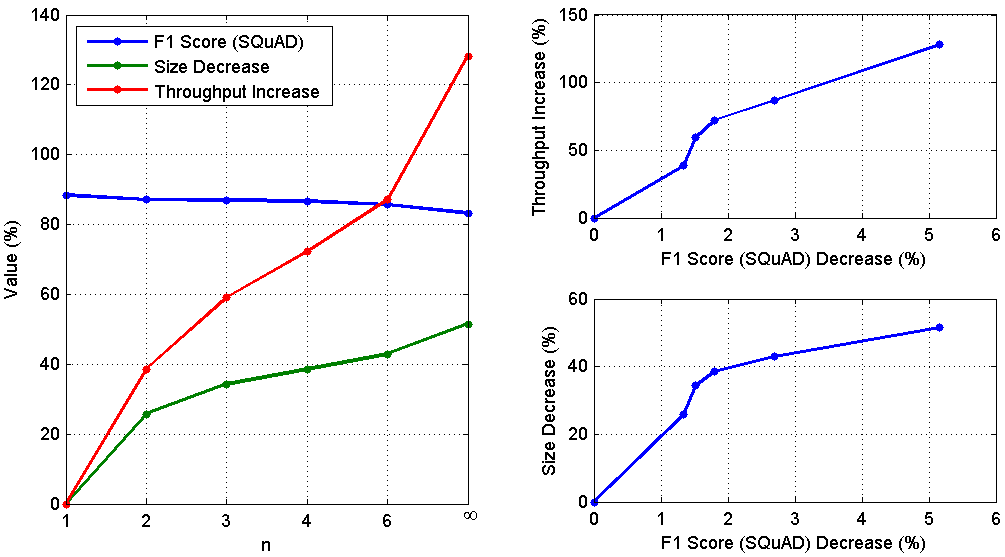}}
    \caption{\label{fig:fine_tuning_results}F1 scores on SQuAD of modified BERT\textsubscript{BASE} networks with different values of $n$ along with the corresponding decrease in size and increase in throughput as a relative percentage to the unmodified ($n$ = 1) network. F1 score, throughput increase and size as a function of $n$ (left). On the right, the trade-off between accuracy (absolute decrease in F1 score) and efficiency in either throughput (top right) or memory footprint (bottom right). Modified networks with $n \in \{2, 3, 4\}$ result in a significant size decrease and throughput increase while retaining a similar F1 score on SQuAD.}
\end{figure}

Using the values in Table \ref{tab:combined_fine_tuning_results_parameters}, Figure \ref{fig:fine_tuning_results} shows the SQuAD F1 scores of the modified BERT\textsubscript{BASE} networks along with their corresponding size decrease and throughput increase relative to the unmodified ($n$ = 1) network. The contribution of intermediate blocks to the final fine-tuning accuracy is not significant but their removal can significantly decrease size and increase speed, allowing trade-offs between accuracy and efficiency. For example, the F1 score of the modified ($n = 3$) network decreases by approximately 1.5\% but it is more than 34\% smaller and 52\% faster.

We  compare our approach against certain knowledge distillation techniques such as DistilBERT, BERT-PKD and BERT-of-Theseus (see Table \ref{tab:combined_fine_tuning_results_parameters}). The SQuAD F1 and EM scores of our modified $n=4$ network is approximately 1\% higher when compared to DistilBERT, while having a similar parameter count. The modified $n=4$ network also performs slightly better than BERT-PKD for the MNLI task.

Finally, since we believe that ALBERT would be complementary to our method and could be used to further improve training speed and decrease model size, we did not perform a direct comparison to it. 

\subsection{Intermediate Layer Removal Prior to Fine-Tuning}
\label{section:appendix_1}

Results in \ref{section:fine_tuning} show that removing intermediate layers have little impact on network accuracy. However, can these intermediate layers be removed \textit{before} the modified network is fine-tuned?

The unmodified BERT\textsubscript{BASE} network is first pre-trained as described in Section \ref{section:convergence_results} and then the intermediate layers are removed. After removal of the intermediate layers, the now modified BERT\textsubscript{BASE} network is fine-tuned on different tasks.

\begin{table}[h]
    \footnotesize
    \begin{center}
        \caption{Results with intermediate blocks removed prior to fine-tuning for modified BERT\textsubscript{BASE} networks and different values of $n$. The F1 and EM scores are shown for SQuAD v1.1, m/mm for MNLI, accuracy for SST-2 and F1 score for QQP.}
        \label{tab:layer_removal_prior_to_fine_tuning_results}
        \begin{tabular}{c|c|c|c|c}
            \hline \hline
            \textbf{Network} & \textbf{SQuAD} & \textbf{MNLI} & \textbf{SST-2} & \textbf{QQP} \\
            \hline
            $n$ = 2 & 84.20/75.63 & 79.32/79.93 & 87.73 & 83.94 \\
            $n$ = 3 & 76.99/67.23 & 74.36/74.13 & 83.03 & 82.59\\
            $n$ = 4 & 72.03/60.80 & 73.55/73.77 & 82.68 & 82.16\\
            $n$ = 6 & 65.45/53.10 & 72.11/72.23 & 82.34 & 82.01 \\
            $n$ = $\infty$ & 56.82/43.66 & 72.55/73.20 & 83.02 & 81.57 \\
            \hline \hline
        \end{tabular}
    \end{center}
\end{table}

Table \ref{tab:layer_removal_prior_to_fine_tuning_results} clearly shows a significant degradation in all fine-tuning tasks for the modified BERT\textsubscript{BASE} networks. For example, in SQuAD v1.1,  the decrease in F1 score for $n$ = 3 when the intermediate layers are removed before fine-tuning is 11.44\% compared with Table \ref{tab:combined_fine_tuning_results_parameters} where the decrease is 1.52\% for $n$ = 3. During pre-training it appears that the intermediate layers are performing important transformations to the representations learned in the self-attention blocks. By removing these layers prior to fine-tuning, the self-attention blocks are unable to compensate during fine-tuning.

\subsection{Dependency Parsing}
We use the analysis methods presented in \cite{clark2019does} to further justify the removal of intermediate layers. In BERT, each individual attention head is known to specialize to particular aspects of syntax and the model's overall knowledge about syntax is distributed across multiple attention heads. Attention-based probing classifiers proposed in \cite{clark2019does} measure the overall ability of the model to learn syntactic information. In \cite{clark2019does}, the probing classifiers are evaluated on the dependency parsing task for the Penn Tree Bank (PTB) data set \cite{10.5555/972470.972475}. These classifiers treat the BERT attention outputs as fixed and train only a small number of parameters. Basically, the classifier produces a probability distribution over each word that indicates how likely each other word in the sentence is its syntactic head. The \texttt{Attn + GloVe} probing classifier substantially outperforms other reference baselines in \cite{clark2019does}. This suggests BERT's attention maps have a fairly thorough representation of English syntax.

We use the same \texttt{Attn + GloVe} probing classifier introduced in \cite{clark2019does} for our analysis on intermediate layers. However, we  evaluate on the Universal Dependency gold standard for English, the English Web Treebank (EWT) data set \cite{silveira2014gold} instead of PTB. Table \ref{tab:fine_tuning_results_probing} shows the Unlabeled Attachment Scores (UAS) on EWT for different values of $n$.

For this particular task it appears that fewer intermediate blocks result in an increase of the UAS score. Therefore, this indicates that removing intermediate blocks has little impact on the model's ability to learn syntactic information.

\begin{table}[h]
    \footnotesize
    \begin{center}
        \caption{Attention-based probing classifier results for unmodified and modified BERT\textsubscript{BASE} networks with different values of $n$.}
        \label{tab:fine_tuning_results_probing}
        \begin{tabular}{c|c}
            \hline \hline
             \textbf{Network} & \textbf{UAS Score} \\
            \hline
            Untrained & 24.20\% \\
            $n$ = 1 & 70.60\% \\
            $n$ = 2 & 69.80\% \\
            $n$ = 3 & 71.50\% \\
            $n$ = 4 & 70.80\% \\
            $n$ = 6 & 71.50\% \\
            $n$ = $\infty$ & 73.00\% \\
            \hline \hline
        \end{tabular}
    \end{center}
\end{table}

\section{Analysis of Intermediate Layers}
In this section, we analyze the similarities of self-attention blocks of unmodified and modified BERT\textsubscript{BASE} networks using centered kernel alignment (CKA) \cite{kornblith2019similarity} to better understand the importance of intermediate layers. The CKA similarity metric has useful properties such as invariance to both orthogonal transformations and isotropic scaling which is important when comparing architecturally similar but not identical networks. We follow the procedure in \cite{kornblith2019similarity} and perform the similarity analysis over multiple pre-training and fine-tuning trials of the unmodified ($n$= 1) and modified BERT\textsubscript{BASE} ($n$=2, 3) networks. A total of 5 trials for each unmodified and modified BERT\textsubscript{BASE} network are pre-trained and then fine-tuned (see sections \ref{section:pre_training} and \ref{section:fine_tuning}).

All BERT\textsubscript{BASE} networks denoted as \texttt{Unmodified} or \texttt{Modified} are first fully pre-trained while networks denoted \texttt{Untrained} are not pre-trained and have their weights initialized randomly. However, \textit{all} networks are fine-tuned using SQuAD v1.1. This training procedure is used so that the similarity analysis is applied to only fine-tuned networks.

After fine-tuning, each BERT\textsubscript{BASE} network is used to encode the same 500 sentences from the WebText \cite{Gokaslan2019OpenWeb} data set to produce output activations from each of the self-attention blocks. These output activations are then used as the input to linear-kernel CKA.

\begin{table}[h]
    \footnotesize
    \begin{center}
        \caption{CKA similarities for multiple pairs of unmodified and modified BERT\textsubscript{BASE} networks fine-tuned using SQuAD. Average linear-kernel CKA similarities along with standard deviation are computed over all 12 self-attention blocks using 5 trials of each network.}
        \label{tab:unmodified_modified_pre_train_squad_similarities}
        \begin{tabular}{c|c}
            \hline \hline
            \textbf{Network Pair} & \textbf{Average CKA Similarity} \\
            \hline
            Untrained vs. n = 1 & 0.17 $\pm$ 0.06 \\
            Untrained vs. n = 2 & 0.22 $\pm$ 0.09 \\
            Untrained vs. n = 3 & 0.21 $\pm$ 0.10 \\
            n = 1 vs. n = 1 & 0.90 $\pm$ 0.08 \\
            n = 1 vs. n = 2 & 0.72 $\pm$ 0.15 \\
            n = 1 vs. n = 3 & 0.72 $\pm$ 0.13 \\
            \hline \hline
        \end{tabular}
    \end{center}
\end{table}

Table \ref{tab:unmodified_modified_pre_train_squad_similarities} shows the average similarities over all 12 self-attention blocks for different pairs of untrained, unmodified and modified BERT\textsubscript{BASE} networks. Similarities between unmodified and modified networks are much higher than similarities with untrained networks. This high similarity translates to a minimal loss in fine-tuning accuracy.

\section{Conclusion}
In this work we proposed a modification to the BERT architecture focusing on reducing the number of intermediate layers in the network. With the modified BERT\textsubscript{BASE} network we show that the network complexity can be significantly decreased while preserving accuracy on fine-tuning tasks. In addition, our analysis with CKA and probing linear classifiers justifies the removal of intermediate layers.

For future work, we plan to apply our modifications to larger models like BERT\textsubscript{LARGE} and GPT to show that the architectural changes can be used more widely. We also intend to integrate other network pruning techniques (e.g., ALBERT) with our modified architecture to further decrease size and increase throughput.

\bibliographystyle{unsrt}  
\bibliography{references}

\end{document}